\documentclass[runningheads]{llncs}
\usepackage{graphicx} 	
\usepackage{amsmath}
\usepackage[utf8]{inputenc} 
\usepackage[T1]{fontenc}    
\usepackage{hyperref}       
\usepackage{url}            
\usepackage{booktabs}       
\usepackage{amsfonts}       
\usepackage{nicefrac}       
\usepackage{microtype}      

\usepackage{amsmath}
\usepackage{graphicx}
\usepackage{xcolor}
\usepackage{etoolbox}
\usepackage[colorinlistoftodos]{todonotes}
\usepackage{makecell}
\usepackage{subcaption}

\usepackage[utf8]{inputenc}
\usepackage{fourier} 
\usepackage{array}
\usepackage{makecell}
\usepackage{amsmath}
\usepackage{ragged2e}

\DeclareMathOperator*{\argmin}{arg\,min}
\usepackage{booktabs}
\newcommand{\head}[1]{\textnormal{\textbf{#1}}}
\newcommand{\normal}[1]{\multicolumn{1}{l}{#1}}

\usepackage{pgfplots}
\pgfplotsset{compat=1.14}
\usepackage{multirow}
\usepackage{float}
\usepackage{gensymb}
\usepackage{blindtext}

\usepackage{fancyhdr}

\begin{document}


\title{Deep Learning in the Wild}

\author{Thilo Stadelmann\inst{1} \and
Mohammadreza Amirian\inst{1,2} \and
Ismail Arabaci\inst{3} \and
Marek Arnold\inst{1,3} \and
Gilbert François Duivesteijn\inst{4} \and
Ismail Elezi\inst{1,5} \and
Melanie Geiger\inst{1,6} \and
Stefan Lörwald\inst{7} \and
Benjamin Bruno Meier\inst{3} \and
Katharina Rombach\inst{1} \and
Lukas Tuggener\inst{1,8}}

\authorrunning{Stadelmann et al.}
\titlerunning{Deep Learning in the Wild}

\institute{ZHAW Datalab \& School of Engineering, Winterthur, Switzerland \and
Institute of Neural Information Processing, Ulm University, Germany \and
ARGUS DATA INSIGHTS Schweiz AG, Zürich, Switzerland \and
Deep Impact AG, Winterthur, Switzerland \and
DAIS, Ca' Foscari University of Venice, Venezia Mestre, Italy \and
Institut d'Informatique, Université de Neuchâtel, Switzerland \and
PricewaterhouseCoopers AG, Zürich, Switzerland \and
IDSIA Dalle Molle Institute for Artificial Intelligence, Manno, Switzerland}

\maketitle

\vspace{-0.5cm}
\begin{abstract}
Deep learning with neural networks is applied by an increasing number of people outside of classic research environments, due to the vast success of the methodology on a wide range of machine perception tasks. While this interest is fueled by beautiful success stories, practical work in deep learning on novel tasks without existing baselines remains challenging. This paper explores the specific challenges arising in the realm of real world tasks, based on case studies from research \& development in conjunction with industry, and extracts lessons learned from them. It thus fills a gap between the publication of latest algorithmic and methodical developments, and the usually omitted nitty-gritty of how to make them work. Specifically, we give insight into deep learning projects on face matching, print media monitoring, industrial quality control, music scanning, strategy game playing, and automated machine learning, thereby providing best practices for deep learning in practice. 
\end{abstract}

\keywords{data availability \and deployment \and loss \& reward shaping \and real world tasks}

\section{Introduction}
\label{sec:introduction}

\setcounter{footnote}{0} 

Measured for example by the interest and participation of industry at the annual NIPS conference\footnote{See \url{https://medium.com/syncedreview/a-statistical-tour-of-nips-2017-438201fb6c8a}.}, it is save to say that deep learning \cite{schmidhuber2015deep} has successfully transitioned from pure research to application \cite{liu2017survey}. Major research challenges still exist, e.g. in the areas of model interpretability \cite{olah2018buildingblocks} and robustness \cite{akhtar2018threat}, or general understanding \cite{shwartz2017opening} and stability \cite{zheng2016improving,irpan2018deeprl} of the learning process, to name a few. Yet, and in addition, another challenge is quickly becoming relevant: in the light of more than $180$ deep learning publications per day in the last year\footnote{Google scholar counts $>68,000$ articles for the year 2017 as of June 11, 2018.}, the growing number of deep learning engineers as well as prospective researchers in the field need to get educated on best practices and what works and what doesn't \emph{``in the wild''}. This information is usually underrepresented in publications of a field that is very competitive and thus striving above all for novelty and benchmark-beating results \cite{olah2017research}. Adding to this fact, with a notable exception \cite{goodfellow2016dlbook}, the field lacks authoritative and detailed textbooks by leading representatives. Learners are thus left with preprints \cite{ng2018yearning,stadelmann2018beyondimagenet}, cookbooks \cite{perez2017playbook}, code\footnote{See e.g. \url{https://modelzoo.co/}.} and older gems \cite{lecun1998efficient,larochelle2009exploring,sutskever2013importance} to find much needed practical advice. 

In this paper, we contribute to closing this gap between cutting edge research and application in the wild by presenting case-based best practices. Based on a number of successful industry-academic research \& development collaborations, we report what specifically enabled success in each case alongside open challenges. The presented findinds (a) come from real-world and business case-backed use cases beyond purely academic competitions; (b) go deliberately beyond what is usually reported in our research papers in terms of tips \& tricks, thus complementing them by the stories behind the scenes; (c) include also what didn't work despite contrary intuition; and (d) have been selected to be transferable as lessons learned to other use cases and application domains. The inteded effect is twofold: more successful applications, and increased applied reasearch in the areas of the remaining challenges.

We organize the main part of this paper by case studies to tell the story behind each undertaking. Per case, we briefly introduce the application as well as the specific (research) challenge behind it; sketch the solution (referring details to elsewhere, as the final model architecture etc. is not the focus of this work); highlight what measures beyond textbook knowledge and published results where necessary to arrive at the solution; and show, wherever possible, examples of the arising difficulties to exemplify the challenges. Section \ref{sec:facematch} introduces a \emph{face matching} application and the amount of surrounding models needed to make it practically applicable. Likewise, Section \ref{sec:mediamonitoring} describes the additional amount of work to deploy a state-of-the-art machine learning system into the wider IT system landscape of an \emph{automated print media monitoring} application. Section \ref{sec:visualquality} discusses interpretability and class imbalance issues when applying deep learning for \emph{images-based industrial quality control}. In Section \ref{sec:musicscanning}, measures to cope with the instability of the training process of a complex model architecture for large-scale \emph{optical music recognition} are presented, and the class imbalance problem has a second appearance. Section \ref{sec:gameplaying} reports on practical ways for deep reinforcement learning in \emph{complex strategy game play} with huge action and state spaces in non-stationary environments. Finally, Section \ref{sec:automl} presents first results on comparing practical \emph{automated machine learning} systems with the scientific state of the art, hinting at the use of simple baseline experiments. Section \ref{sec:conclusions} summarizes the lessons learned and gives an outlook on future work on deep learning in practice.

\section{Face matching}
\label{sec:facematch}

Designing, training and testing deep learning models for application in face recognition comes with all the well known challenges like choosing the architecture, setting hyperparameters, creating a representative training/dev/test dataset, preventing bias or overfitting of the trained model, and more. Anyway, very good results have been reported in the literature \cite{vggface2015,schroff2015facenet,vgg2face2017}. Although the challenges in lab conditions are not to be taken lightly, a new set of difficulties emerges when deploying these models in a real product. Specifically, during development, it is known what to expect as input in the controlled environment. When the models are integrated in a product that is used ``in the wild'', however, all kinds of input can reach the system, making it hard to maintain a consistent and reliable prediction. In this section, we report on approaches to deal with related challenges in developing an actual face-ID verification product.

\begin{figure}[t]
	\begin{center}
    \includegraphics[width=1.0\textwidth]{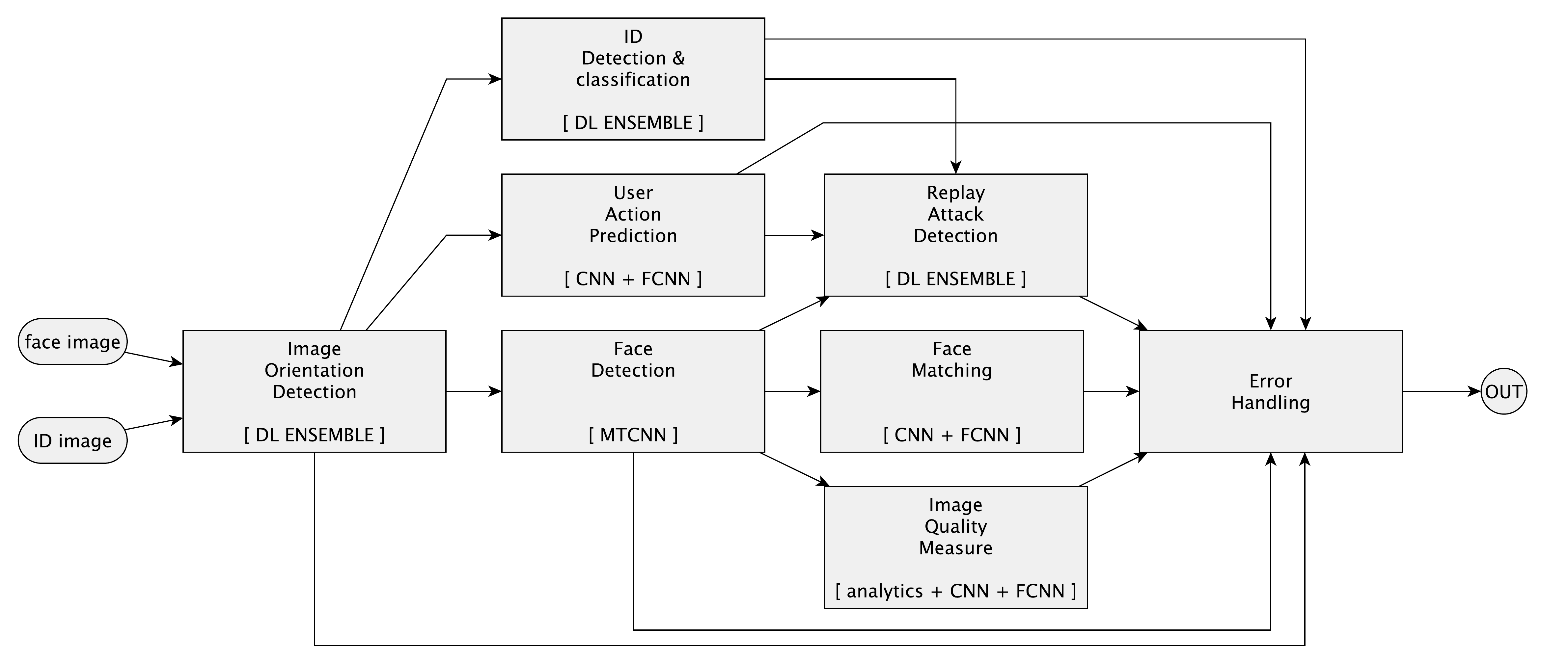}
    \caption{Schematic representation of a face matching application with ID detection, anti-spoofing and image quality assessment. For any pair of input images (selfie and ID document), the output is the match probability and type of ID document, if no anomaly or attack has been detected. Note that all boxes contain at least one or several deep learning (DL) models with many different (convolutional) architectures.}
    \label{fig:gfd1}
    \end{center}
    \vspace{-0.8cm}
\end{figure}
    
Although the core functionality of such a product is to quantify the match between a person's face and the photo on the given ID, more functionality is needed to make the system perform its task well, most of it hidden from the user. Thus, in addition to the actual face matching module, the final system contains at least the following machine learnable modules (see Figure \ref{fig:gfd1}):
\begin{description}
	\item [Image orientation detection] When a user takes a photo of the ID on a flat surface using a mobile phone, in many cases the image orientation is random. A deep learning method is applied to predict the orientation angle, used to rotate the image in the correct orientation.
	\item [Image quality assessment] consists of an ensemble of analytical functions and deep learning models to test if the photo quality is sufficient for a reliable match. It also guides the user to improve the picture taking process in case of bad quality.
	\item [User action prediction] uses deep learning to predict the action performed by the user to guide the system's workflow, e.g. making a selfie, presenting an ID or if the user is doing something wrong during the sequence.
	\item [Anti-Spoofing] is an essential module that uses various methods to detect if a person is showing his ``real'' face or tries to fool the system with a photo, video or mask. It consists of an ensemble of deep learning models.
\end{description}

For a commercial face-ID product, the anti-spoofing module is both most crucial for success, and technically most challenging; thus, the following discussion will focus on anti-spoofing in practice. Face matching and recognition systems are vulnerable to spoofing attacks made by non-real faces, because they are not per se able to detect whether or not a face is ``live'' or ``not-live'', given only a single image as input in the worst case. If control over this input is out of the system's reach e.g. for product management reasons, it is then easy to fool the face matching system by showing a photo of a face from screen or print on paper, a video or even a mask. To guard against such spoofing, a secure system needs to be able to do live-ness detection. We'd like to highlight the methods we use for this task, in order to show the additional complexity of applying face recognition in a production environment over lab conditions.

\begin{figure}[t]
    \begin{center}
    \begin{tabular}{c c c c c c}
    \includegraphics[width=0.14\textwidth]{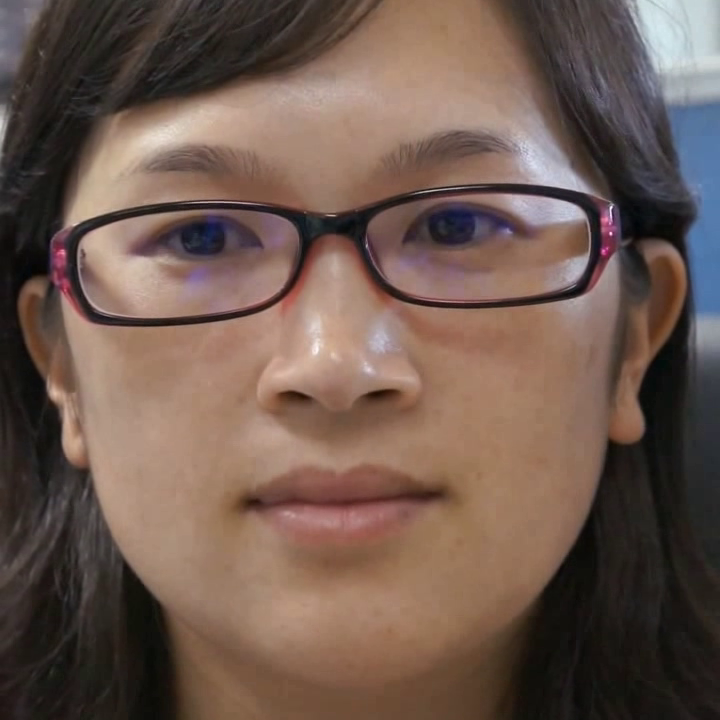} &
    \includegraphics[width=0.14\textwidth]{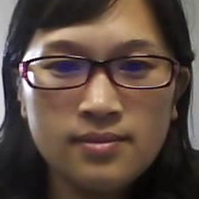} &
    \includegraphics[width=0.14\textwidth]{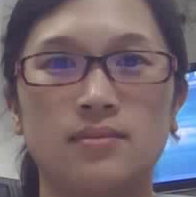} &
    \includegraphics[width=0.14\textwidth]{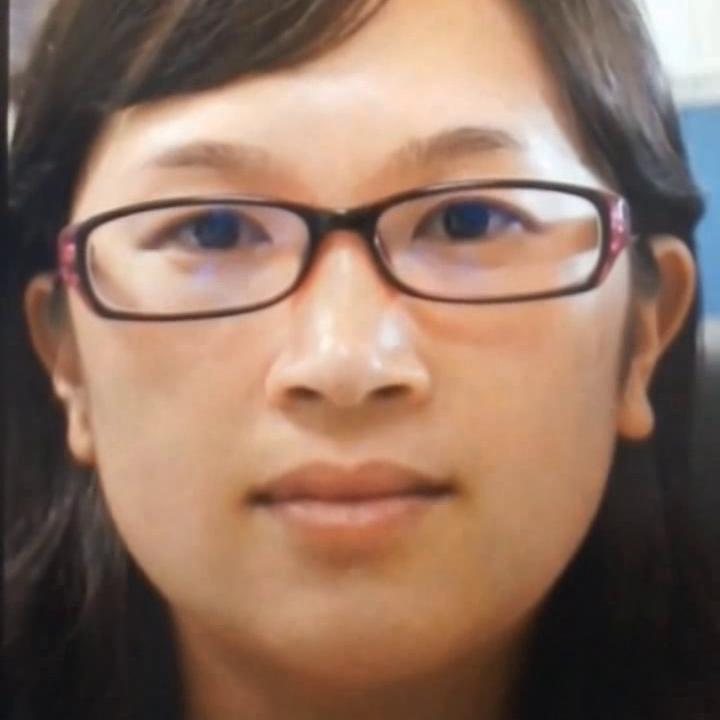} &
    \includegraphics[width=0.14\textwidth]{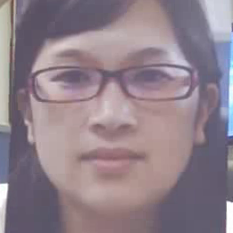} &
    \includegraphics[width=0.14\textwidth]{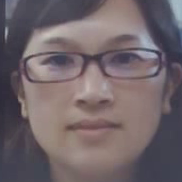} 
    \end{tabular}
    \caption{Samples from the CASIA dataset \cite{DBLP:conf/icb/ZhangYLLYL12}, where photo 1, 2, and 3 on the left hand side show a real face, photo 4 shows a replay attack from a digital screen, and photos 5 and 6 show replay attacks from print.}
    \label{fig:gfd2}
    \end{center}
    \vspace{-1.0cm}
\end{figure}
    
One of the key features of spoofed images is that they usually can be detected because of degraded image quality: when taking a photo of a photo, the quality deteriorates. However, with high quality cameras in modern mobile phones, looking at image quality only is not sufficient in the real world. How then can a spoof detector be designed that approves a real face from a low quality grainy underexposed photo taken by an old $640 \times 480$ web cam, and rejects a replay attack using a photo from a retina display in front of a $4$K video camera (compare Figure \ref{fig:gfd2})? 
    
Most of the many spoofing detection methods proposed in the literature use hand crafted features, followed by shallow learning techniques, e.g. SVM \cite{DBLP:journals/tip/GalballyMF14,DBLP:conf/icb/MaattaHP11,li_wang_tan_jain_2004}. These techniques mainly focus on texture differences between real and spoofed images, differences in color space \cite{DBLP:conf/icip/BoulkenafetKH15}, Fourier spectra \cite{li_wang_tan_jain_2004}, or optical flow maps \cite{bao_li_li_jiang_2009}. In more recent work, deep learning methods have been introduced \cite{DBLP:conf/icb/AtoumLJ017,DBLP:journals/corr/YangLL14,DBLP:conf/acpr/XuLD15,DBLP:conf/ipta/LiFBXLH16}. Most methods have in common that they attempt to be a one-size-fits-all solution, classifying all incoming cases with one method. This might be facilitated by the available datasets: to develop and evaluate anti-spoofing tools, amongst others CASIA \cite{DBLP:conf/icb/ZhangYLLYL12}, MSU-USSA \cite{DBLP:journals/tifs/PatelHJ16}, and the Replay Attack Database \cite{DBLP:conf/biosig/ChingovskaAM12} exist. Although these datasets are challenging, they turn out to be too easy compared to the input in a production environment. 

The main differences between real cases and training examples from these benchmark databases are that the latter ones have been created with a low variety of hardware devices and only use few different locations and light conditions. Moreover, the quality of images throughout the training sets is quite consistent, which does not reflect real input. In contrast, the images that the system receives ``in the wild'' have the most wide range of possible used hardware and environmental conditions, making the anticipation of new cases difficult. Designing a single system that can classify all such cases with high accuracy seems therefore unrealistic.
	
We thus create an ensemble of experts, forming a final verdict from $3$ independent predictions: the first method consists of $2$ patch-based CNNs, one for low resolution images, the other one for high resolution images. They operate on fixed-size tiles from the unscaled input image using a sliding window. This technique proves to be effective for low and high quality input. The second method uses over $20$ image quality measures as features combined with a classifier. This method is still very effective when the input quality is low. The third method uses a RNN with LSTM cells to conduct a joint prediction over multiple frames (if available). It is effective in discriminating micro movements of a real face against (simple) translations and rotations of a fake face, e.g. from a photo on paper or screen. All methods return a real vs. fake probability. The outputs of all $3$ methods are fed as input features to the final decision tree classifier. This ensemble of deep learning models is experimentally determined to be much more accurate than using any known method individually. 

Note that as attackers are inventive and come up with new ways to fool the system quickly, it is important to update the models with new data quickly and regularly.

\section{Print media monitoring}
\label{sec:mediamonitoring}

\begin{figure}[t]
    \centering
    \begin{subfigure}[b]{0.28\columnwidth}
        \centering
        \includegraphics[width=\columnwidth]{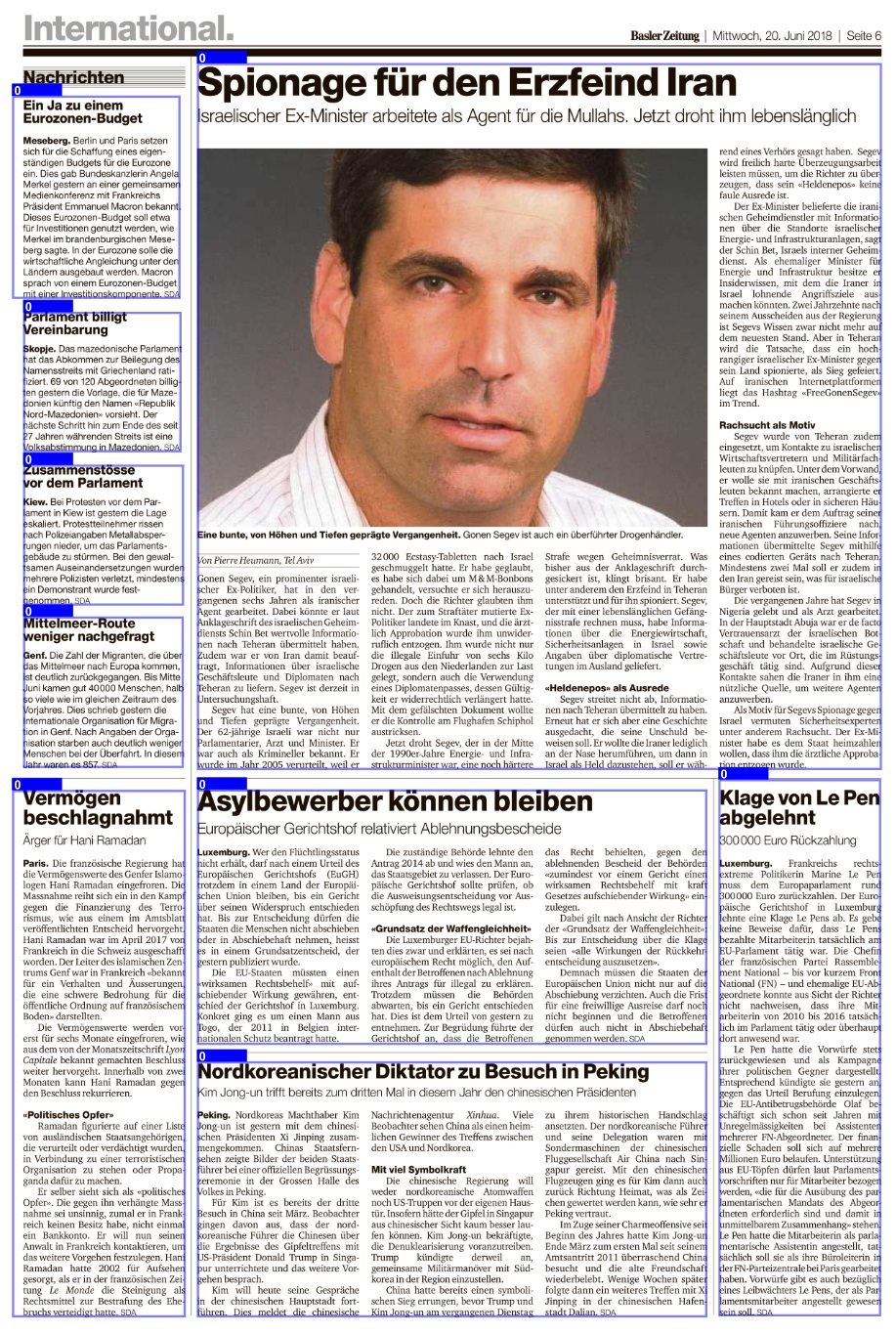}
        \caption{}
        \label{fig:panoptes_goodex}
    \end{subfigure}
    \begin{subfigure}[b]{0.28\columnwidth}
        \centering
        \includegraphics[width=\columnwidth]{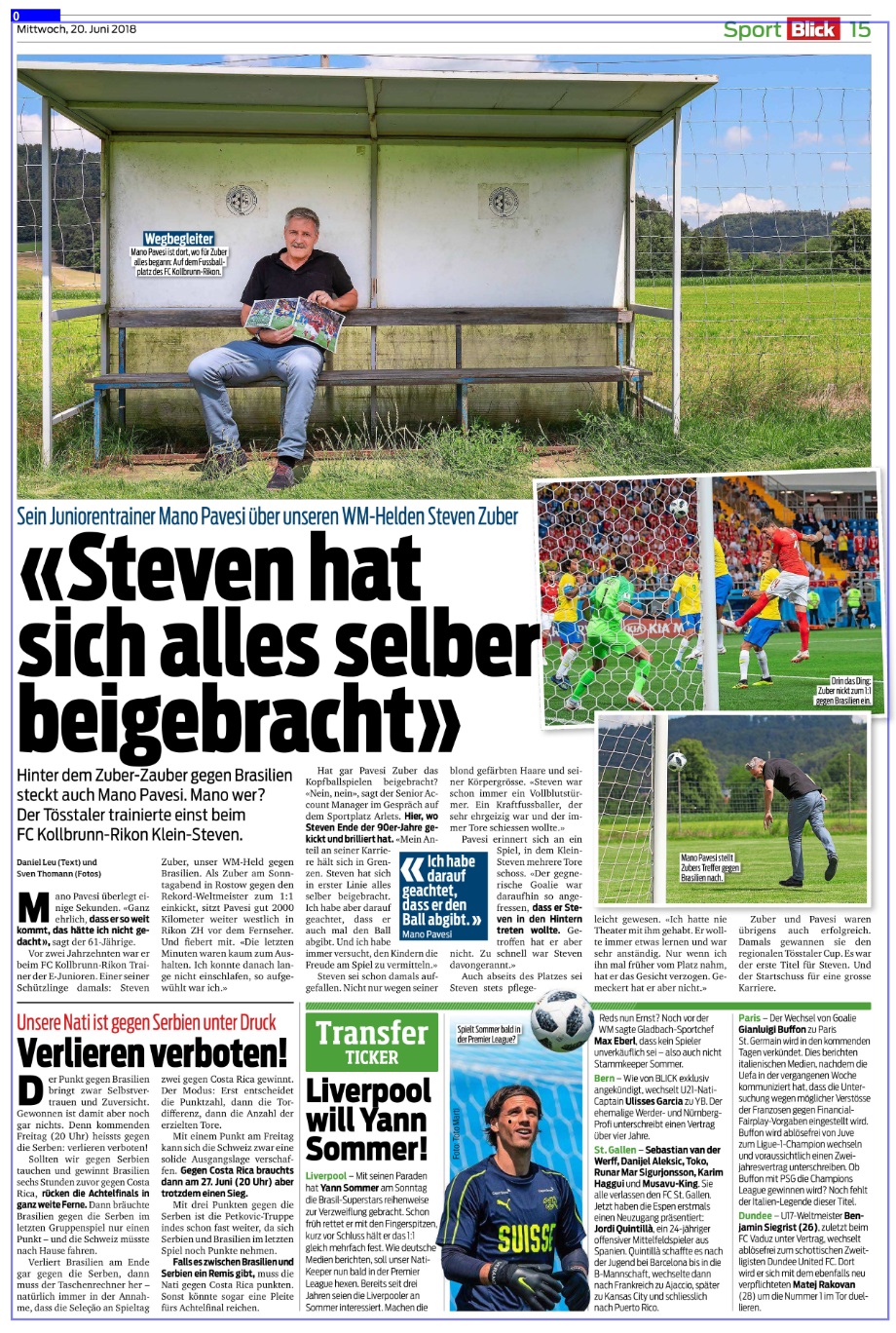}
        \caption{}
        \label{fig:panoptes_badex}
    \end{subfigure}
    \begin{subfigure}[b]{0.27\columnwidth}
        \centering
        \includegraphics[width=\columnwidth]{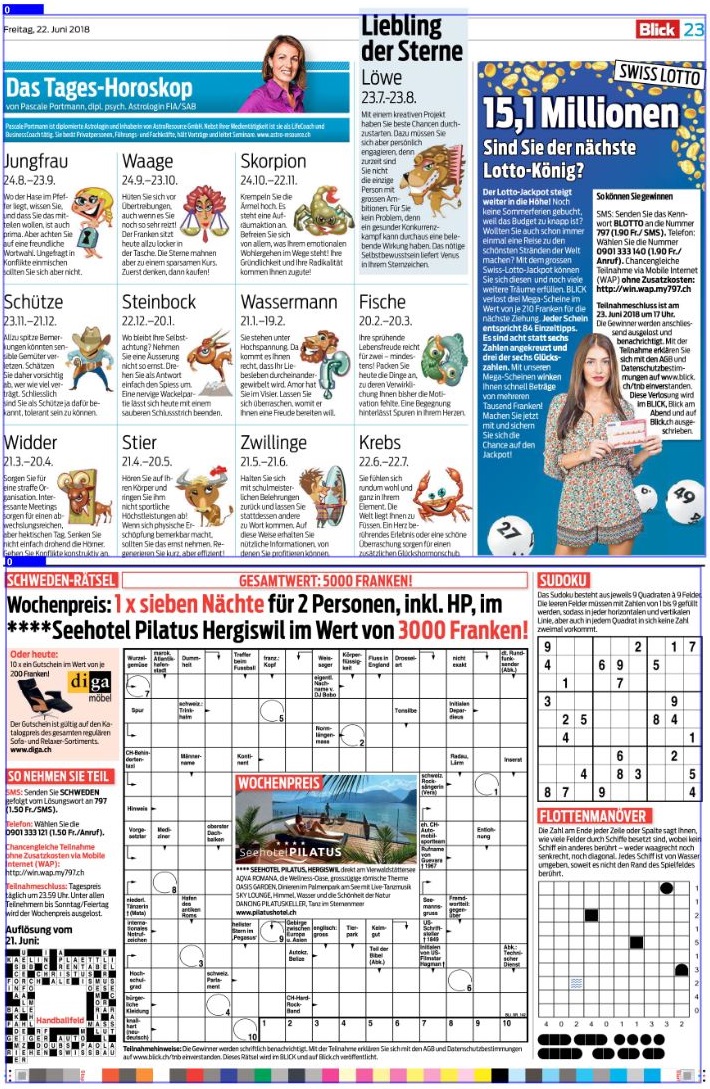}
        \caption{}
        \label{fig:panoptes_special}
    \end{subfigure}    
  \caption{Good (a) and bad (b) segmentations (blue lines denote crop marks) for realistic pages, depending on the freedom in the layout. Image (c) shows a non-article page that is excluded from automatic segmentation.}
  \label{fig:panoptes_ex}
  \vspace{-0.5cm}
\end{figure}

Content-based print media monitoring serves the task of delivering cropped digital articles from printed newspapers to customers based on their pre-formulated information need (e.g., articles about their own coverage in the media). For this form of article-based information retrieval, it is necessary to segment tens of thousands of newspaper pages into articles daily. We successfully developed neural network-based models to learn how to segment pages into their constituting articles and described their details elsewhere \cite{stadelmann2018beyondimagenet,meier2017fully} (see example results in Figure \ref{fig:panoptes_ex}a-b). In this section, we present challenges faced and learnings gained from integrating a respective model into a production environment with strict performance and reliability requirements.

\vspace{-0.2cm}
\paragraph{\textbf{Exclusion of non-article pages}} A common problem in print segmentation are special pages that contain content that doesn't represent articles in the common sense, for example classified ads, reader's letters, TV program, share prices, or sports results (see Figure \ref{fig:panoptes_special}). Segmentation rules for such pages can be complicated, subjective, and provide little value for general use cases. We thus utilize a random forest-based classifier on handcrafted features to detect such content and avoid feeding respective pages to the general segmentation system to save compute time.

\vspace{-0.2cm}
\paragraph{\textbf{Model management}} One advantage of an existing manual segmentation pipeline is the abundance of high quality, labeled training data being produced daily. To utilize this constant flow of data, we have started implementing an online learning system \cite{shalev2012online} where results of the automatic segmentation can be corrected within the regular workflow of the segmentation process and fed back to the system as training data.

After training, an important business decision is the final configuration of a model, e.g. determining a good threshold for cuts to weigh between precision and recall, or the decision on how many different models should be used for the production system. We determined experimentally that it is more effective to train different models for different publishers: the same publisher often uses a similar layout even for different newspapers and magazines, while differences between publishers are considerable. To simplify the management of these different models, they are decoupled from the code. This is helpful for rapid development and experimentation.

\begin{figure}[t]
  \begin{center}
    \includegraphics[width=1.0\textwidth]{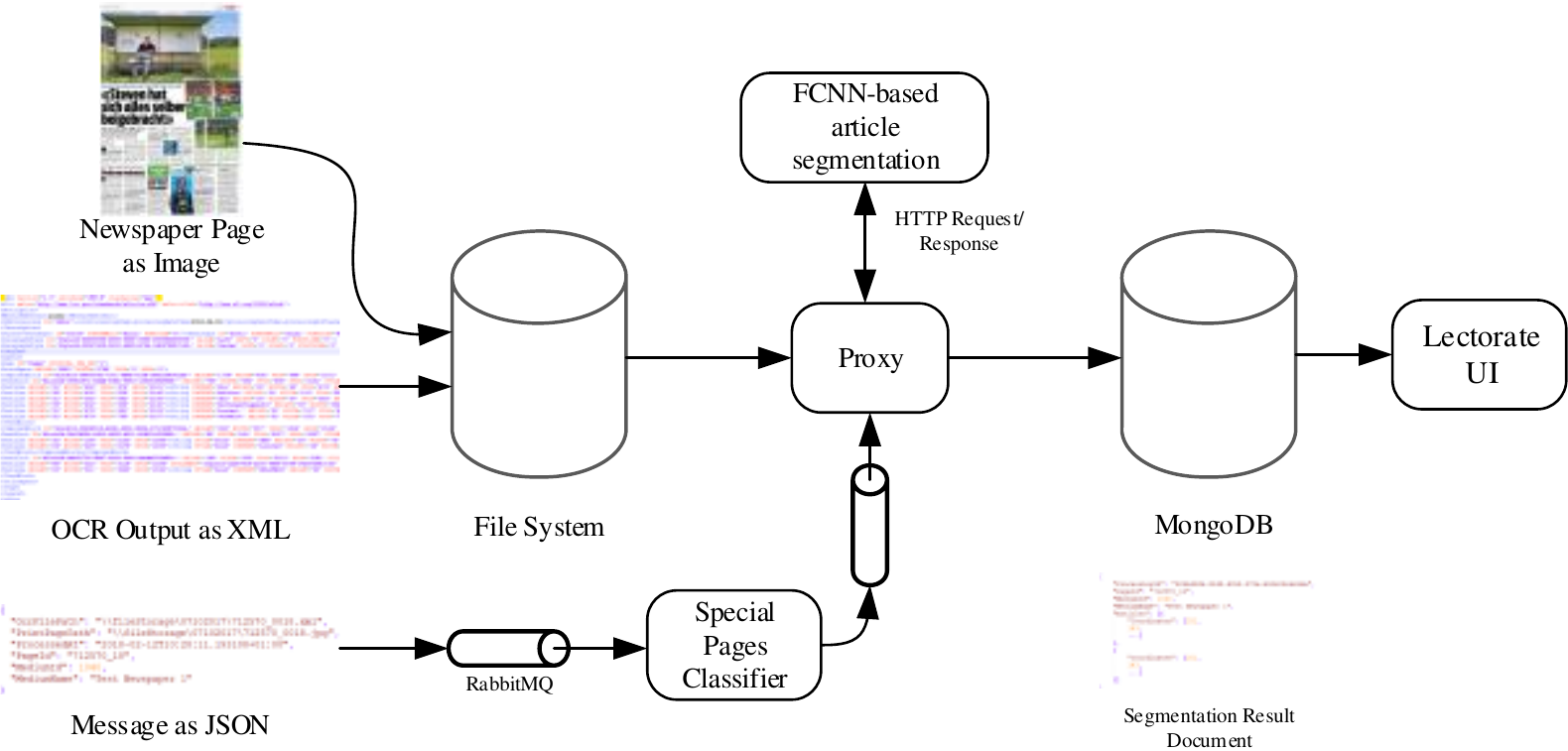}
    \caption{Architecture of the overall pipeline: the actual model is encapsulated in the ``FCNN-based article segmentation'' block. Several other systems are required to warrant full functionality: (a) the \emph{Proxy} is responsible to control data input and output from the segmentation model; (b) \emph{RabbitMQ} controls the workflow as a message broker; (c) \emph{MongoDB} stores all segmentation results and metrics; (d) the \emph{Lectorate} UI visualizes results for human assessment and is used to create training data.}
    \label{fig:panoptes_arch}
  \end{center}
  \vspace{-0.8cm}
\end{figure}

\vspace{-0.2cm}
\paragraph{\textbf{Technological integration}} For smooth development and operation of the neural network application we have chosen to use a containerized microservices architecture \cite{dragoni2017microservices} utilizing Docker \cite{xu2017performance} and RabbitMQ \cite{DBLP:journals/corr/JohnL17}. This decoupled architecture (see Figure \ref{fig:panoptes_arch}) brings several benefits especially for machine learning applications: (a) a \emph{separation of concerns} between research, ops and engineering tasks; (b) \emph{decoupling of models/data from code}, allowing for rapid experimentation and high flexibility when deploying the individual components of the system. This is further improved by a modern devops pipeline consisting of continuous integration (CI), continuous deployment (CD), and automated testing; (c) \emph{infrastructure flexibility}, as the entire pipeline can be deployed to an on-premise data center or in the cloud with little effort. Furthermore, the use of Nvidia-docker \cite{xu2017performance} allows to utilize GPU-computing easily on any infrastructure; (d) precise \emph{controlling and monitoring} of every component in the system is made easy by data streams that enable the injection and extraction of data such as streaming event arguments, log files, and metrics at any stage of the pipeline; and (e) easy \emph{scaling} of the various components to fit different use cases (e.g. training, testing, experimenting, production). Every scenario requires a certain configuration of the system for optimal performance and resource utilization.

\section{Visual quality control}
\label{sec:visualquality}

Manual inspection of medical products for in-body use like balloon catheters is time-consuming, tiring and thus error-prone. A semi-automatic solution with high precision is thus sought. In this section, we present a case study of deep learning for visual quality control of industrial products. While this seems to be a standard use case for a CNN-based approach, the task differs in several interesting respects from standard image classification settings:

\setcounter{table}{4}
\renewcommand{\tablename}{Fig.}

\begin{table}[t]
     \begin{center}
     \begin{tabular}{ c  c  c  c}
     \includegraphics[width=0.2\textwidth]{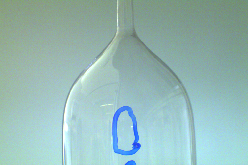} &              \includegraphics[width=0.2\textwidth]{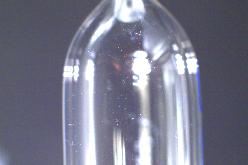} &  \includegraphics[width=0.2\textwidth]{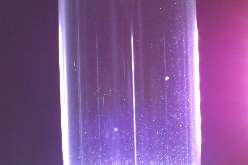} & \includegraphics[width=0.2\textwidth]{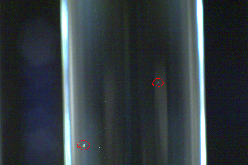}  \\ 
      \end{tabular}
      \caption{Balloon catheter images taken under different optical conditions, exposing (left to right) high reflections, low defect visibility, strong artifacts, and a good setup.}
      \label{fig:Mo_1}
      \end{center}
      \vspace{-1.2cm}
\end{table}

\vspace{-0.2cm}
\paragraph{\textbf{Data collection}} and labeling are one the most critical issues in most practical applications. Detectable defects in our case appear as small anomalies on the surface of transparent balloon catheters, such as scratches, inclusions or bubbles. Recognizing such defects on a thin, transparent and reflecting plastic surface is visually challenging even for expert operators that sometimes refer to a microscope to manually identify the defects. Thus, approx. $50$\% of a $2$-year project duration was used on finding and verifying the optimal optical settings for image acquisition. Figure \ref{fig:Mo_1} depicts the results of different optical configurations for such photo shootings. Finally, operators have to be trained to produce consistent labels usable for a machine learning system. In our experience, the labeling quality rises if all involved parties have a basic understanding of the methods. This helps considerably to avoid errors like e.g. only to label a defect on the first image of a series of shots while rotating a balloon: while this is perfectly reasonable from a human perspective (once spotted, the human easily tracks the defect while the balloon moves), it is a no-go for the episodic application of a CNN.

\vspace{-0.2cm}
\paragraph{\textbf{Network and training design}} for practical applications experiences challenges such as class imbalance, small data regimes, and use case-specific learning targets apart from standard classification settings, making non-standard loss functions necessary (see also Section \ref{sec:musicscanning}). For instance, in the current application, we are looking for relatively small defects on technical images. Therefore, architectures proposed for large-scale natural image classification such as AlexNet \cite{krizhevsky2012imagenet}, GoogLeNet \cite{szegedy2015going}, ResNet \cite{he2016deep} and modern variants are not necessarily successful, and respective architectures have to be adapted to learn the relevant task. Potential solutions for the class imbalance problem are for example:
\begin{itemize}
  \item Down-sampling the majority class 
  \item Up-sampling the minority class via image augmentation \cite{DBLP:conf/cvpr/CiresanMS12}
  \item Using pre-trained networks and applying transfer learning \cite{pan2010survey} 
  \item Increasing the weight of the minority class in the optimization loss  \cite{buda2017systematic}
  \item Generating synthetic data for the minority class using SMOTE \cite{chawla2002smote} or GANs \cite{goodfellow2014explaining}
\end{itemize}
Selecting a suitable data augmentation approach according for the task is a necessity for its success. For instance, in the present case, axial scratches are more important than radial ones, as they can lead to a tearing of the balloon and its subsequent potentially lethal remaining in a patient's body. Thus, using $90\degree$ rotation for data augmentation could be fatal. Information like this is only gained in close collaboration with domain experts.     

\begin{table}[t]
     \begin{center}
     \begin{tabular}{l | c c c c}
     \toprule
      & Image & Feature response & Image & Feature response \\ \midrule
      Negative & \includegraphics[width=0.14\textwidth]{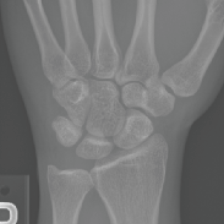} &                \includegraphics[width=0.14\textwidth]{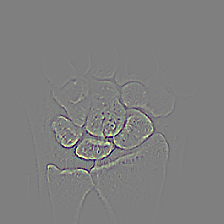} & \includegraphics[width=0.14\textwidth]{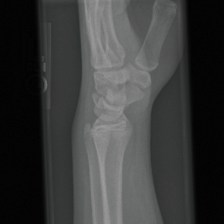} & \includegraphics[width=0.14\textwidth]{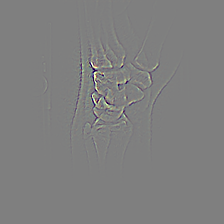} \\ \midrule
      Positive & \includegraphics[width=0.14\textwidth]{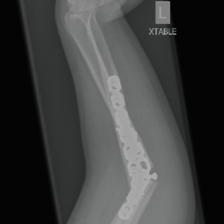} & \includegraphics[width=0.14\textwidth]{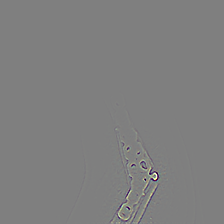} & \includegraphics[width=0.14\textwidth]       {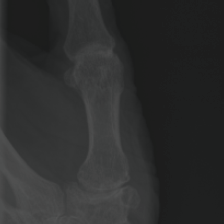} & \includegraphics[width=0.14\textwidth]{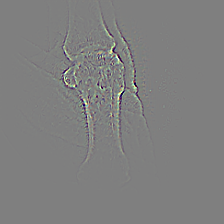} \\ 
      \toprule
      \end{tabular}
	  \caption{Visualizing VGG19 feature responses: the first row contains two negative examples (healthy patient) and the second row positives (containing anomalies). All depicted samples are correctly classified.}
	  \label{fig:Mo_2}
      \end{center}
      \vspace{-1.2cm}
\end{table}

\begin{table}[t]
     \begin{center}
     \begin{tabular}{l | c c c c}
     \toprule
      & Original & Adversarial & Original & Adversarial \\ \midrule
      Image & \includegraphics[width=0.14\textwidth]{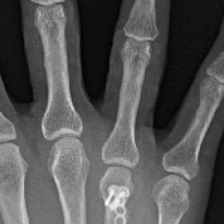} &                \includegraphics[width=0.14\textwidth]{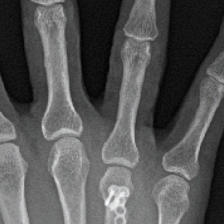} &      \includegraphics[width=0.14\textwidth]{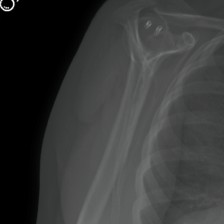} & \includegraphics[width=0.14\textwidth]{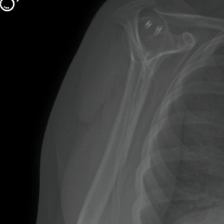} \\ \midrule
     Feature response & \includegraphics[width=0.14\textwidth]{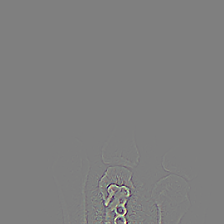} &                \includegraphics[width=0.14\textwidth]{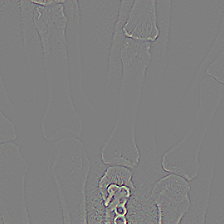} & \includegraphics[width=0.14\textwidth]       {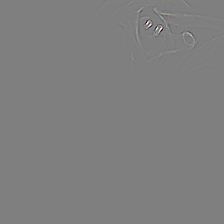} & \includegraphics[width=0.14\textwidth]{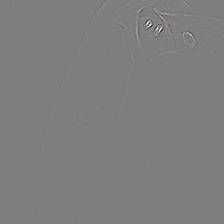} \\ \midrule
     Local spatial entropy & \includegraphics[width=0.14\textwidth]{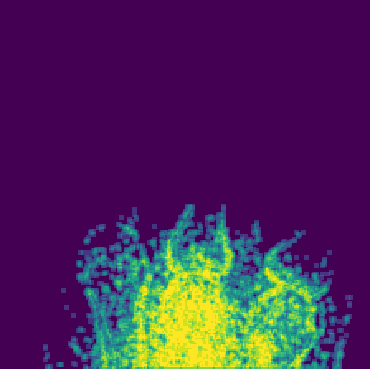} &                \includegraphics[width=0.14\textwidth]{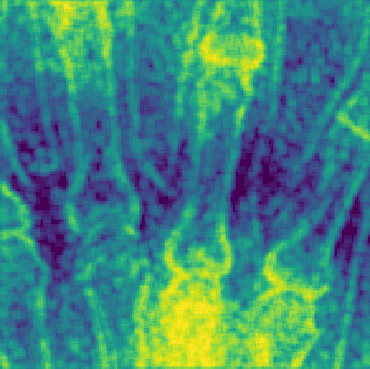} & \includegraphics[width=0.14\textwidth]       {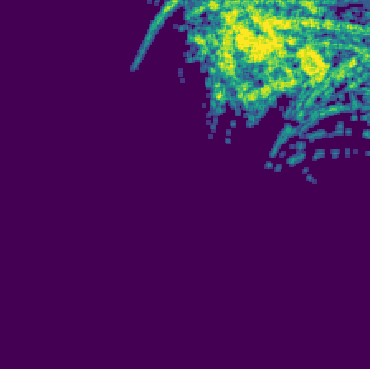} & \includegraphics[width=0.14\textwidth]{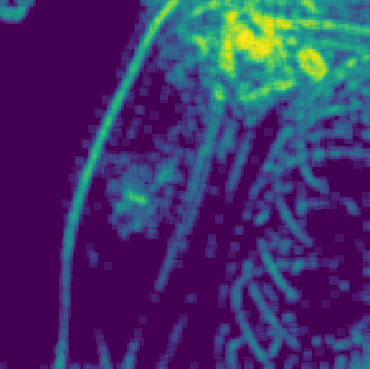} \\ \midrule
      Predicted class & Positive & Negative & Positive & Negative \\
      \toprule
      \end{tabular}
	  \caption{Input, feature response and local spatial entropy for clean and adversarial images, respectively. We used VGG19 to estimate predictions and the Fast Gradient Sign Attack (FGSM) method \cite{goodfellow2014explaining} to compute the adversarial perturbation.}
	  \label{fig:Mo_3}
      \end{center}
      \vspace{-1.2cm}
\end{table}

\renewcommand{\tablename}{Table}
\setcounter{table}{0}

\vspace{-0.2cm}
\paragraph{\textbf{Interpretability}} of models received considerable attention recently, spurring hopes both of users for transparent decisions, and of experts for ``debugging'' the learning process. The latter might lead for instance to improved learning from few labeled examples through semantic understanding of the middle layers and intermediate representations in a network. Figure \ref{fig:Mo_2} illustrates some human-interpretable representations of the inner workings of a CNN on the recently published MUsculoskeletal RAdiographs (MURA) dataset \cite{rajpurkar2017mura} that we use here as a proxy for the balloon dataset. We used guided-backpropagation \cite{springenberg2014striving} and a standard VGG19 network \cite{simonyan2014very} to visualize the feature responses, i.e. the part of the X-ray image on which the network focuses for its decision on ``defect'' (e.g., broken bone, foreign object) or ``ok'' (natural and healthy body part). It can be seen that the network mostly decides based on joints and detected defects, strengthening trust in its usefulness. We described elsewhere \cite{amirian2018trace} that this visualization can be extended to an automatic defense against adversarial attacks \cite{goodfellow2014explaining} on deployed neural networks by thresholding the local spatial entropy \cite{chanwimaluang2003efficient} of the feature response. As Figure \ref{fig:Mo_3} depicts, the focus of a model under attack widens considerably, suggesting that it ``doesn't know where to look'' anymore.

\section{Music scanning}
\label{sec:musicscanning}

Optical music recognition (OMR) \cite{DBLP:journals/ijmir/RebeloFPMGC12} is the process of translating an image of a page of sheet music into a machine-readable structured format like MusicXML. Existing products exhibit a symbol recognition error rate that is an order of magnitude too high for automatic transcription under professional standards, but don't leverage deep learning computer vision capabilities yet. In this section, we therefore report on the implementation of a deep learning approach to detect and classify all musical symbols on a full page of written music in one go, and integrate our model into the open source system Audiveris\footnote{See \url{http://audiveris.org}.} for the semantic reconstruction of the music. This enables products like digital music stands based on active sheets, as most of todays music is stored in image-based PDF files or on paper.

We highlight four typical issues when applying deep learning techniques to practical OMR: (a) the absence of a comprehensive dataset; (b) the extreme class imbalance present in written music with respect to symbols; (c) the issues of state-of-the-art object detectors with music notation (many tiny and compound symbols on large images); and (d) the transfer from synthetic data to real world examples. 

\setcounter{figure}{7}

\begin{figure}[t]
  \centering
  \includegraphics[width=1\textwidth]{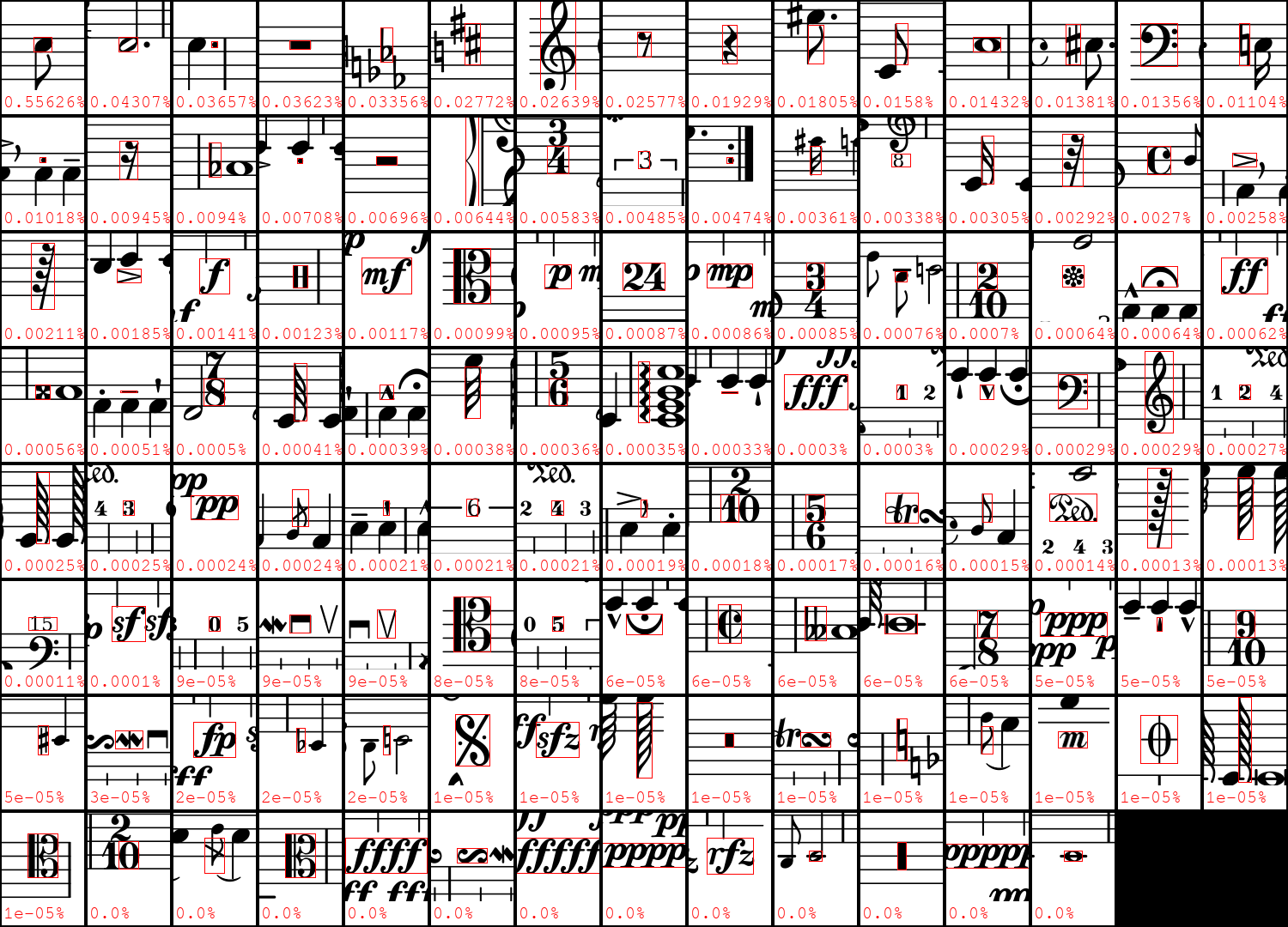}
  \caption{Symbol classes in \emph{DeepScores} with their relative frequencies (red) in the dataset.}
  \label{fig:deepscores_imbalance}
  \vspace{-0.7cm}
\end{figure}

\vspace{-0.2cm}
\paragraph{\textbf{Synthesizing training data}} The notorious data hunger of deep learning has lead to a strong dependence of results on large, well annotated datasets, such as ImageNet \cite{ILSVRC15} or PASCAL VOC \cite{DBLP:journals/ijcv/EveringhamGWWZ10}. For music object recognition, no such dataset has been readily available. Since labeling data by hand is no feasible option, we put a one-year effort in synthesizing realistic (i.e., semantically and syntactically correct music notation) data and the corresponding labeling from renderings of publicly available MusicXML files and recently open sourced the resulting \emph{DeepScores} dataset \cite{tuggener2018deepscores}. 

\vspace{-0.2cm}
\paragraph{\textbf{Dealing with imbalanced data}} While typical academic training datasets are nicely balanced \cite{ILSVRC15,DBLP:journals/ijcv/EveringhamGWWZ10}, this is rarely the case in datasets sourced from real world tasks. Music notation (and therefore \emph{DeepScores}) shows an extreme class imbalance (see Figure \ref{fig:deepscores_imbalance}). For example, the most common class (note head black) contains more than $55$\% of the symbols in the entire dataset, and the top $10$ classes contain more than $85$\% of the symbols. At the other extreme, there is a class which is present only once in the entire dataset, making its detection by pattern recognition methods nearly impossible (a ``black swan'' is no pattern). However, symbols that are rare are often of high importance in the specific pieces of music where they appear, so simply ignoring the rare symbols in the training data is not an option. A common way to address such  imbalance is the use of a weighted loss function, as described in Section \ref{sec:visualquality}. 

This is not enough in our case: first, the imbalance is so extreme that naively reweighing loss components leads to numerical instability; second, the signal of these rare symbols is so sparse that it will get lost in the noise of the stochastic gradient descent method \cite{DBLP:conference/ismir/tuggener}, as many symbols will only be present in a tiny fraction of the mini batches. Our current answer to this problem is \emph{data synthesis} \cite{ng2018yearning}, using a three-fold approach to synthesize image patches with rare symbols (cp. Figure \ref{fig:deepscores_imbalance}): (a) we locate rare symbols which are present at least $300$ times in the dataset, and crop the parts containing those symbols including their local context (other symbols, staff lines etc.); (b) for rarer symbols, we locate a semantically similar but more common symbol in the dataset (based on some expert-devised notion of symbol similarity), replace this common symbol with the rare symbol and add the resulting page to the dataset. This way, synthesized sheets still have semantic sense, and the network can learn from syntactically correct context symbols. We then crop patches around the rare symbols similar to the previous approach; (c) for rare symbols without similar common symbols, we automatically ``compose''  music containing those symbols. 

Then, during training, we augment each input page in a mini batch with $12$ randomly selected synthesized crops of rare symbols (of size $130 \times 80$ pixels) by putting them in the margins at the top of the page. This way, that the neural network (on expectation) does not need to wait for more than $10$ iterations to see every class which is present in the dataset. Preliminary results show improvement, though more investigation is needed: overfitting on extreme rare symbols is still likely, and questions remain regarding how to integrate the concept of patches (in the margins) with the idea of a full page classifier that considers all context.

\begin{figure}[t]
  \centering
  \includegraphics[width=1\textwidth]{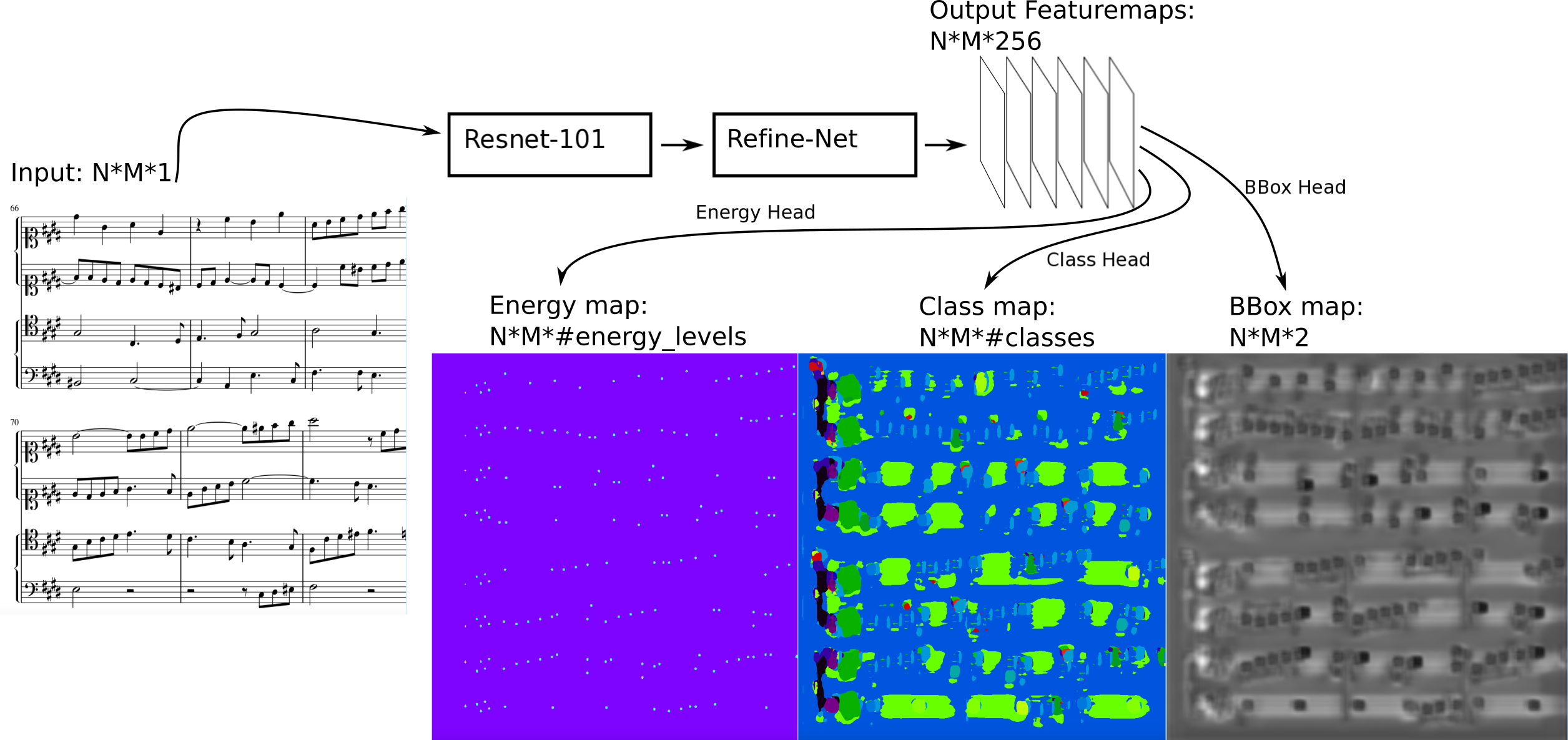}
  \caption{Schematic of the Deep Watershed Detector model with three distinct output heads. $N$ and $M$ are the height and width of the input image, $\mathrm{\#classes}$ denotes the number of symbols and $\mathrm{\#energy\_levels}$ is a hyperparameter of the system.}
  \label{fig:dwd}
  \vspace{-0.5cm}
\end{figure}

\vspace{-0.2cm}
\paragraph{\textbf{Enabling \& stabilizing training}} We initially used state-of-the-art object detection models like Faster R-CNN \cite{DBLP:conf/nips/RenHGS15} to attempt detection and classification of musical symbols on \emph{DeepScores}. These algorithms are designed to work well on the prevalent datasets that are characterized by containing low-resolution images with a few big objects. In contrast, \emph{DeepScores} consists of high resolution musical sheets containing hundreds of very small objects, amounting to a very different problem \cite{tuggener2018deepscores}. This disconnect lead to very poor out-of-the-box performance of said systems.

Region proposal-based systems scale badly with the number of objects present on a given image, by design. Hence, we designed the \emph{Deep Watershed Detector} as an entirely new object detection system based on the deep watershed transform \cite{DBLP:conf/cvpr/BaiU17} and described it in detail elsewhere \cite{DBLP:conference/ismir/tuggener}. It detects raw musical symbols (e.g., not a compound note, but note head, stem and flag individually) in their context with a full sheet music page as input. As depicted in Figure \ref{fig:dwd}, the underlying neural network architecture has three output heads on the last layer, each pertaining to a separate (pixel wise) task: (a) predicting the underlying symbol's class; (b) predicting the energy level (i.e., the degree of belonging of a given pixel location to an object center, also called "objectness"); and (c) predicting the bounding box of the object. 

Initially, the training was unstable, and we observed that the network did not learn well if it was directly trained on the combined weighted loss. Therefore, we now train the network on each of the three tasks separately. We further observed that while the network gets trained on the bounding box prediction and classification, the energy level predictions get worse. To avoid this, the network is fine-tuned only for the energy level loss after being trained on all three tasks. Finally, the network is retrained on the combined task (the sum of all three losses, normalized by their respective running means) for a few thousand iterations, giving excellent results on common symbols.

\begin{figure}[t]
  \centering
  \includegraphics[width=0.75\textwidth]{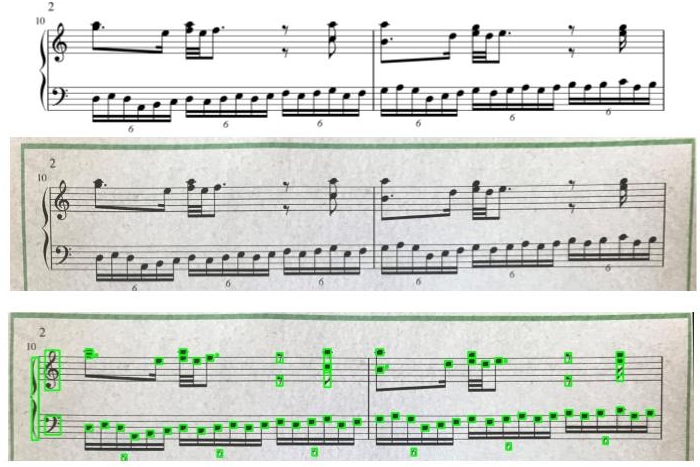}
  \caption{Top: part of a synthesized image from \emph{DeepScores}; middle: the same part, printed on old paper and photographed using a cell phone; bottom: the same image, automatically retrofitted (based on the dark green lines) to the original image coordinates for ground truth matching (ground truth overlayed in neon green boxes).}
  \label{fig:realworld}
  \vspace{-0.7cm}
\end{figure}

\vspace{-0.2cm}
\paragraph{\textbf{Generalizing to real-world data}} The basic assumption in machine learning for training and test data to stem from the same distribution is often violated in field applications. In the present case, domain adaptation is crucial: our training set consists of synthetic sheets created by LilyPond scripts \cite{tuggener2018deepscores}, while the final product will work on scans or photographs of printed sheet music. These test pictures can have a wide variety of impairments, such as bad printer quality, torn or stained paper etc. While some work has been published on the topic of \emph{domain transfer} \cite{DBLP:conf/iccv/GebruHF17}, the results are non-satisfactory. The core idea to address this problem here is transfer learning \cite{DBLP:conf/nips/YosinskiCBL14}: the neural network shall learn the core task of the full complexity of music notation from the synthetic dataset (symbols in context due to full page input), and use a much smaller dataset to adapt to the real world distributions of lighting, printing and defect.

We construct this post-training dataset by carefully choosing several hundred representative musical sheets, printing them with different types of printers on different types of paper, and finally scanning or photographing them. We then use the \texttt{BFMatcher} function from OpenCV to align these images with the original musical sheets to use all the ground truth annotation of the original musical sheet for the real-world images (see Figure \ref{fig:realworld}). This way, we get annotated real-looking images ``for free'' that have much closer statistics to real-world images than images from \emph{DeepScores}. With careful tuning of the hyperparameters (especially the regularization coefficient), we get promising - but not perfect - results during the inference stage.

\section{Game playing}
\label{sec:gameplaying}

In this case study, deep reinforcement learning (DRL) is applied to an agent in a multi-player  business simulation video game with steadily increasing complexity, comparable to StarCraft or SimCity. The agent is expected to compete with human players in this environment, i.e. to continuously adapt its strategy to challenge evolving opponents. Thus, the agent is required to mimic somewhat general intelligent behavior by transferring knowledge to an increasingly complex environment and adapting its behavior and strategies in a non-stationary, multi-agent environment with large action and state spaces. DRL is a general paradigm, theoretically able to learn any complex task in (almost) any environment. In this section, we share our experiences with applying DRL to the above described competitive environment. Specifically, the performance of a value-based algorithm using Deep Q-Networks (DQN) \cite{mnih2013playing} is compared to a policy gradient method called PPO \cite{schulman2017proximal}.

\vspace{-0.2cm}
\paragraph{\textbf{Dealing with competitive environments}} In recent years, astounding results have been achieved by applying DRL in gaming environments. Examples are Atari games \cite{mnih2013playing} and AlphaGo \cite{silver2016mastering}, where agents learn human or superhuman performance purely from scratch. In both examples, the environments are either stationary or, if an evolving opponent is present, it did not act simultaneously in the environment; instead, actions were taken in turns. In our environment, multiple evolving players act simultaneously, making changes to the environment that can not be explained solely based on changes in the agent's own policy. Thus, the environment is perceived as non-stationary from the agent's perspective, resulting in stability issues in RL \cite{lowe2017multi}. Another source of complexity in our setting is a huge action and state space (see below). In our experiments, we observed that DQN got problems learning successful control policies as soon as the environment became more complex in this respect, even without non-stationarity induced by opponents. On the other hand, PPO's performance is generally less sensitive to increasing state and action spaces. The impact of non-stationarity to these algorithms is subject of ongoing work.

\vspace{-0.2cm}
\paragraph{\textbf{Reward shaping}} An obvious rewarding choice is the current score of the game (or its gain). Yet, in the given environment, scoring and thus any reward based on it is sparse, since it is dependent on a long sequence of correct actions on the operational, tactical and strategic level. As any rollout of the agent without scoring is not contributing to any gain in knowledge, the learning curve is flat initially. To avoid this initial phase of no information gain, intermediate rewards are given to individual actions, leading to faster learning progress in both DQN and PPO. 

Additionally, it is not sufficient for the agent to find a control policy eventually, but it is crucial to find a good policy \emph{quickly}, as training times are anyhow very long. Usually, comparable agents for learning complex behaviors in competitive environments are trained using self-play \cite{bansal2017emergent}, i.e., the agents are always trained with ``equally good'' competitors to be able to succeed eventually. In our setting, self play is not a straightforward first option, for several reasons: first, to jump-start learning, it is easier in our setting to play without an opponent first and only learn the art of competition later when a stable ability to act is reached; second, different from other settings, our agents should be entertaining to human opponents, not necessarily winning. It is thus not desirable to learn completely new strategies that are successful yet frustrating to human opponents. Therefore, we will investigate self-play only after stable initializations from (scripted) human opponents on different levels.

\begin{figure}[t]
  \centering
  \includegraphics[width=0.75\textwidth]{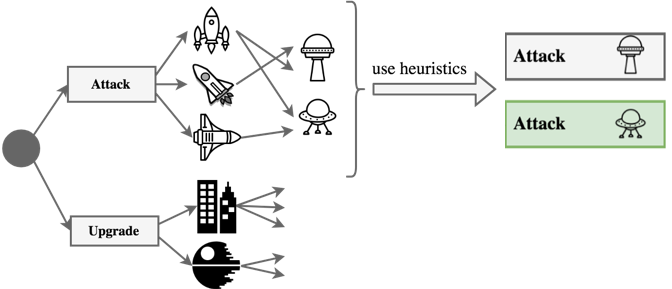}
  \caption{Heuristic encoding of actions to prevent combinatorial explosion.}
  \label{fig:actionchoice}
  \vspace{-0.7cm}
\end{figure}

\vspace{-0.2cm}
\paragraph{\textbf{Complex state and action spaces}} Taking the screen frame (i.e., pixels) as input to the control policy is not applicable in our case. First, the policy's input needs to be independent of rendering and thus of hardware, game settings, game version etc. Furthermore, a current frame does not satisfy the Markov property, since attributes like ``I own item $x$'' are not necessarily visible in it. Instead, some attributes need to be concluded from past experiences. Thus, the state space needs to be encoded into sufficient features, a task we approach with manual pre-engineering.

Next, a post-engineering approach helps in decreasing the learning time in case of DQN by removing unnecessary actions from consideration as follows: in principal, RL algorithms explore any theoretically possible state-action pair in the environment, i.e., any mathematically possible decision in the Markov Decision Process (MDP). In our environment, the available actions are dependent on the currently available in-game resources of the player, i.e., on the current state. Thus, exploring currently impossible regions in the action space is not efficient and is thus prevented by a post-engineered decision logic built to block these actions from being selected. This reduces the size of the action space per time stamp considerably. These rules where crucial in producing first satisfying learning results in our environment using DQN in a stationary setting of the game. However, when training the agent with PPO, hand-engineered rules where not necessary for proper learning. 

The major problem however is the huge action and state space, as it leads to ever longer training times and thus long development cycles. It results from the fact that one single action in our environment might consist of a sequence of sub-decisions. Think e.g. of an action called ``attack'' in the game of StarCraft, answering the question of WHAT to do (see Figure \ref{fig:actionchoice}). It is incompletely defined as long as it does not state WHICH opponent is to be attack using WHICH unit. In other words, each action itself requires a number of different decisions, chosen from different subcategories. To avoid the combinatorial explosion of all possible completely defined actions, we perform another post-processing on the resource management: WHICH unit to choose on WHICH type of enemy, for example, is hard-coded into heuristic rules.

This case study is work in progress, but what becomes evident already is that the combination of the complexity of the task (i.e., acting simultaneously on the operational, tactical and strategic level with exponentially increasing time horizons, as well as a huge state and action space) and the non-stationary environment prevent successful end-to-end learning as in ``Pong from pixels''\footnote{Compare \url{http://karpathy.github.io/2016/05/31/rl/}.}. Rather, it takes manual pre- and post-engineering to arrive at a first agent that learns, and it does so better with policy-based rather than DQN-based algorithms. A next step will explore an explicitly hierarchical learner to cope with the combinatorial explosion of the action space on the three time scales (operational/tactical/strategic) without using hard-coded rules, but instead factorizing the action space into subcategories.

\section{Automated machine learning}
\label{sec:automl}

One of the challenging tasks in applying machine learning successfully is to select a suitable algorithm and set of hyperparameters for a given dataset. Recent research in automated machine learning \cite{feurer2015efficient,Olson2016EvoBio} and respective academic challenges \cite{Guyon:IJCNN:2015} accurately aimed at finding a solution to this problem for sets of practically relevant use cases. The respective Combined Algorithm Selection and Hyperparameter (CASH) optimization problem is defined as finding the best algorithm $A^*$ and set of hyperparameters $\lambda_*$ with respect to an arbitrary cross-validation loss $\mathcal{L}$ as follows:   

\vspace{-0.5cm}   
\begin{align*}
 A^*, \lambda_* = \argmin_{A \in \mathcal{A}, \lambda \in \Lambda_{A} } \frac{1}{K} \sum_{i=1}^{K} \mathcal{L}(A_\lambda,D_{train}^{(i)},D_{valid}^{(i)}) 
\end{align*}

where $\mathcal{A}$ is a set of algorithms, $\Lambda_A$ the set of hyperparameters per algorithm $A$ (together they form the hypothesis space), $K$ is the number of cross validation folds and $D$ are datasets. In this section, we compare two methods from the scientific state-of-the-art (one uses Bayesian optimization, the other genetic programming) with a commercial automated machine learning prototype based on random search.

\vspace{-0.2cm}
\paragraph{\textbf{Scientific state-of-the-art}} Auto-sklearn \cite{feurer2015efficient} is the most successful automated machine learning framework in past competitions \cite{Guyon:AutoML:2016}. The algorithm starts with extracting meta-features from the given dataset and finds models which perform well on similar datasets (according to the meta-features) in a fixed pool of stored successful machine learning endeavors. Auto-sklearn then performs meta-learning by initializing a set of model candidates with the model and hyperparameter choices of $k$ nearest neighbors in dataset space; subsequently, it optimizes their hyperparameters and feature preprocessing pipeline using Bayesian optimization. Finally, an ensemble of the optimized models is build using a greedy search. On the other side, Tree-based Pipeline Optimization Tool (TPOT) \cite{Olson2016EvoBio} is toolbox based on genetic programming. The algorithm starts with random initial configurations including feature preprocessing, feature selection and a supervised classifier. At every step, the top $20$\% best models are retained and randomly modified to generate offspring. The offspring competes with the parent, and winning models proceed to the next iteration of the algorithm.

\vspace{-0.2cm}
\paragraph{\textbf{Commercial prototype}} The Data Science Machine (DSM) is currently used inhouse for data science projects by a business partner. It uses random sampling of the solution space for optimization. Machine learning algorithms in this system are leveraged from Microsoft Azure, scikit-learn and can be user-enhanced. DSM can be deployed in the cloud, on-premise, as well as standalone. The pipeline of DSM includes data preparation, feature reduction, automatic model optimization, evaluation and final ensemble creation. The question is: can it prevail against much more sophisticated systems even at this early stage of development?

\begin{table}[t]
    \resizebox{\textwidth}{!}{\begin{tabular}{l l  l| c c c c c c}
		\toprule[1.5pt]
		& & & \multicolumn{2}{c}{\head{Auto-Sklearn}} &
		\multicolumn{2}{c}{\head{TPOT}} & \multicolumn{2}{c}{\head{DSM}}\\
		\textbf{Dataset} & \textbf{Task} & \textbf{Metric} & \normal{\head{Validation}} &
		\textbf{Test} & \textbf{Validation} & \textbf{Test} & \textbf{Validation} &  \textbf{Test} \\
		\cmidrule(lr){1-9}     
Cadata        & Regression & Coefficient Of Determination &        0.7913        &        0.7801        &     \textbf{0.8245}        &        \textbf{0.8017}        &        0.7078        &        0.7119          \\ 
Christine     & Binary Classification & Balanced Accuracy Score &        0.7380        &        0.7405        &            \textbf{0.7435}        &        \textbf{0.7454}        &          0.7362        &        0.7146         \\ 
Digits        & Multiclass Classification & Balanced Accuracy Score &        \textbf{0.9560}        &        \textbf{0.9556}        &      0.9500        &        0.9458        &        0.8900        &        0.8751        \\ 
Fabert        & Multiclass Classification & Accuracy Score &        \textbf{0.7245}        &        \textbf{0.7193}       &        0.7172        &        0.7006        &        0.7112        &        0.6942        \\ 
Helena        & Multiclass Classification & Balanced Accuracy Score &       \textbf{0.3404}      &       \textbf{0.3434}       &        0.2654        &        0.2667       &        0.2085        &        0.2103        \\ 
Jasmine       & Binary Classification & Balanced Accuracy Score &        0.7987        &        0.8348       &        \textbf{0.8188}        &        0.8281      &        0.8020        &       \textbf{ 0.8371}      \\ 
Madeline      & Binary Classification & Balanced Accuracy Score &        \textbf{0.8917}        &       \textbf{ 0.8769}        &          0.8885        &        0.8620       &        0.7707        &        0.7686      \\ 
Philippine    & Binary Classification & Balanced Accuracy Score &        0.7787        &        0.7486        &        \textbf{0.7839}        &        \textbf{0.7646}       &        0.7581        &        0.7406      \\
Sylvine       & Binary Classification & Balanced Accuracy Score &        0.9414        &        0.9454        &        \textbf{0.9512}        &        \textbf{0.9493}        &        0.9414        &        0.9233        \\ 
Volkert       & Multiclass Classification & Accuracy Score &        \textbf{0.7174}        &        \textbf{0.7101}       &        0.6429        &        0.6327        &        0.5220        &        0.5153        \\            	
    	\cmidrule(lr){1-9} 
    	\multicolumn{2}{l}{\textbf{Average Performance}} & &  \textbf{0.7678}        &        \textbf{0.7654}       &        0.7586        &        0.7497       &        0.7048        &        0.6991     \\
		
		\bottomrule[1.5pt]
	\end{tabular}}
    \caption{Comparison of different automated machine learning algorithms.}
    \label{table:ada}
	\vspace{-1.0cm}    
\end{table}

\vspace{-0.2cm}
\paragraph{\textbf{Evaluation}} is performed using the protocol of the AutoML challenge \cite{Guyon:IJCNN:2015} for comparability, confined to a subset of ten datasets that is processable for the current DSM prototype (i.e., non-sparse, non-big). It spans the tasks of regression, binary and multi-class classification. For applicability, we constrain the time budget of the searches by the required time for DSM to train $100$ models using random algorithm selection. A performance comparison is given in Table \ref{table:ada}, suggesting that Bayesian optimization and genetic programming are superior to random search. However, random parameter search lead to reasonably good models and useful results as well (also in commercial practice). This suggests room for improvement in actual meta-\emph{learning}.

\section{Conclusions}
\label{sec:conclusions}

Does deep learning work in the wild, in business and industry? In the light of the presented case studies, a better questions is: \emph{what does it take to make it work?} Apparently, the challenges are different compared to academic competitions: instead of a given task and known (but still arbitrarily challenging) environment, given by data and evaluation metric, real-world applications are characterized by (a) data quality and quantity issues; and (b) unprecedented (thus: unclear) learning targets. This reflects the different nature of the problems: competitions provide a controlled but unexplored environment to facilitate the discovery of new methods; real-world tasks on the other hand build on the knowledge of a zoo of methods (network architectures, training methods) to solve a specific, yet still unspecified (in formal terms) task, thereby enhancing the method zoo in return in case of success. The following lessons learned can be drawn from our six case studies (section numbers given in parentheses refer to respective details):
\begin{description}
	\item [Data] acquisition usually needs much more time than expected (\ref{sec:visualquality}), yet is the basis for all subsequent success (\ref{sec:musicscanning}). Class imbalance and covariate shift are usual (\ref{sec:facematch},\ref{sec:visualquality},\ref{sec:musicscanning}).
    \item [Understanding] of what has been learned and how decisions emerge help both the user and the developer of neural networks to build trust and improve quality (\ref{sec:visualquality},\ref{sec:musicscanning}). Operators and business owners need a basic understanding of used methods to produce usable ground truth and provide relevant subject matter expertise (\ref{sec:visualquality}).
    \item [Deployment] should include online learning (\ref{sec:mediamonitoring}) and might involve the buildup of up to dozens of other machine learning models (\ref{sec:facematch}, \ref{sec:mediamonitoring}) to flank the original core part. 
    \item [Loss/reward shaping] is usually necessary to enable learning of very complex target functions in the first place (\ref{sec:musicscanning},\ref{sec:gameplaying}). This includes encoding expert knowledge manually into the model architecture or training setup (\ref{sec:visualquality}, \ref{sec:gameplaying}), and handling special cases separately (\ref{sec:mediamonitoring}) using some automatic pre-classification.
    \item [Simple baselines] do a good job in determining the feasibility as well as the potential of the task at hand when final datasets or novel methods are not yet seen (\ref{sec:visualquality}, \ref{sec:automl}). Increasing the complexity of methods and (toy-)tasks in small increments helps monitoring progress, which is important to effectively debug failure cases (\ref{sec:gameplaying}).
    \item [Specialized models] for identifiable sub-problems increase the accuracy in production systems over all-in-one solutions (\ref{sec:facematch},\ref{sec:mediamonitoring}), and ensembles of experts help where no single method reaches adequate performance (\ref{sec:facematch}).
\end{description}

Best practices are straightforward to extract on the general level (``plan enough resources for data acquisition''), yet quickly get very specific when broken down to technicalities (``prefer policy-based RL given that \dots''). An overarching scheme seems to be that the challenges in real-world tasks need similar amounts of creativity and knowledge to get solved as fundamental research tasks, suggesting they need similar development methodologies on top of proper engineering and business planning.

We identified specific areas for future applied research: (a) \emph{anti-spoofing} for face verification; (b) the \emph{class imbalance} problem in OMR; and (c) the slow learning and poor performance of \emph{RL agents} in non-stationary environments with large action and state spaces. The latter is partially addressed by new challenges like Dota 2\footnote{See e.g. \url{https://blog.openai.com/dota-2/}.}, Pommerman or VizDoom\footnote{See \url{https://www.pommerman.com/competitions} and \url{http://vizdoom.cs.put.edu.pl}.}, but for example doesn't address hierarchical actions. Generally, future work should include (d) making deep learning more \emph{sample efficient} to cope with smaller training sets (e.g. by one-shot learning, data or label generation \cite{DBLP:journals/icpr/elezi}, or architecture learning); (e) finding \emph{suitable architectures} and \emph{loss designs} to cope with the complexity of real-world tasks; and (f) improving the \emph{stability} of training and \emph{robustness} of predictions along with (d) the \emph{interpretability} of neural nets.


\vspace{-0.2cm}
\paragraph{\textbf{Acknowledgements}}
We are grateful for the invitation by the ANNPR chairs and the support of our business partners in Innosuisse grants 17719.1 ``PANOPTES'', 17963.1 ``DeepScore'', 25256.1 ``Libra'', 25335.1 ``FarmAI'', 25948.1 ``Ada'' and 26025.1 ``QualitAI''.

\bibliography{annpr-invited}

\begin{thebibliography}{10}
\providecommand{\url}[1]{\texttt{#1}}
\providecommand{\urlprefix}{URL }
\providecommand{\doi}[1]{https://doi.org/#1}

\bibitem{akhtar2018threat}
Akhtar, N., Mian, A.: Threat of adversarial attacks on deep learning in
  computer vision: A survey. arXiv preprint arXiv:1801.00553  (2018)

\bibitem{amirian2018trace}
Amirian, M., Schwenker, F., Stadelmann, T.: Trace and detect adversarial
  attacks on {CNN}s using feature response maps. In: ANNPR (2018)

\bibitem{DBLP:conf/icb/AtoumLJ017}
Atoum, Y., Liu, Y., Jourabloo, A., Liu, X.: Face anti-spoofing using patch and
  depth-based {CNN}s. In: IEEE Int. Joint Conference on Biometrics (IJCB)
  (2017)

\bibitem{DBLP:conf/cvpr/BaiU17}
Bai, M., Urtasun, R.: Deep watershed transform for instance segmentation. In:
  CVPR (2017)

\bibitem{bansal2017emergent}
Bansal, T., Pachocki, J., Sidor, S., Sutskever, I., Mordatch, I.: Emergent
  complexity via multi-agent competition. arXiv preprint arXiv:1710.03748
  (2017)

\bibitem{bao_li_li_jiang_2009}
Bao, W., Li, H., Li, N., Jiang, W.: A liveness detection method for face
  recognition based on optical flow field. Int. Conference on Image Analysis
  and Signal Processing  (2009)

\bibitem{DBLP:conf/icip/BoulkenafetKH15}
Boulkenafet, Z., Komulainen, J., Hadid, A.: Face anti-spoofing based on color
  texture analysis. In: Int. Conference on Image Processing (ICIP) (2015)

\bibitem{buda2017systematic}
Buda, M., Maki, A., Mazurowski, M.A.: A systematic study of the class imbalance
  problem in convolutional neural networks. arXiv preprint arXiv:1710.05381
  (2017)

\bibitem{vgg2face2017}
Cao, Q., Shen, L., Xie, W., Parkhi, O.M., Zisserman, A.: {VGGFace2}: {A}
  dataset for recognising faces across pose and age. arXiv preprint
  arXiv:1710.08092  (2017)

\bibitem{chanwimaluang2003efficient}
Chanwimaluang, T., Fan, G.: An efficient blood vessel detection algorithm for
  retinal images using local entropy thresholding. Int. Symposium on Circuits
  and Systems (ISCAS)  \textbf{5} (2003)

\bibitem{chawla2002smote}
Chawla, N.V., Bowyer, K.W., Hall, L.O., Kegelmeyer, W.P.: {SMOTE}: synthetic
  minority over-sampling technique. Journal of artificial intelligence research
   \textbf{16},  321--357 (2002)

\bibitem{DBLP:conf/biosig/ChingovskaAM12}
Chingovska, I., Anjos, A., Marcel, S.: On the effectiveness of local binary
  patterns in face anti-spoofing. In: BIOSIG (2012)

\bibitem{DBLP:conf/cvpr/CiresanMS12}
Ciresan, D.C., Meier, U., Schmidhuber, J.: Multi-column deep neural networks
  for image classification. In: CVPR (2012)

\bibitem{dragoni2017microservices}
Dragoni, N., Lanese, I., Larsen, S.T., Mazzara, M., Mustafin, R., Safina, L.:
  Microservices: How to make your application scale. In: International Andrei
  Ershov Memorial Conference on Perspectives of System Informatics. Springer
  (2017)

\bibitem{DBLP:journals/icpr/elezi}
Elezi, I., Torcinovich, A., Vascon, S., Pelillo, M.: Transductive label
  augmentation for improved deep network learning. In: ICPR (2018)

\bibitem{DBLP:journals/ijcv/EveringhamGWWZ10}
Everingham, M., Gool, L.J.V., Williams, C.K.I., Winn, J.M., Zisserman, A.: The
  {PASCAL} visual object classes {(VOC)} challenge. Int. Journal of Computer
  Vision  \textbf{88}(2),  303--338 (2010)

\bibitem{feurer2015efficient}
Feurer, M., Klein, A., Eggensperger, K., Springenberg, J., Blum, M., Hutter,
  F.: Efficient and robust automated machine learning. In: NIPS (2015)

\bibitem{DBLP:journals/tip/GalballyMF14}
Galbally, J., Marcel, S., Fi{\'{e}}rrez, J.: Image quality assessment for fake
  biometric detection: Application to iris, fingerprint, and face recognition.
  {IEEE} Trans. Image Processing  \textbf{23}(2),  710--724 (2014)

\bibitem{DBLP:conf/iccv/GebruHF17}
Gebru, T., Hoffman, J., Fei{-}Fei, L.: Fine-grained recognition in the wild:
  {A} multi-task domain adaptation approach. In: ICCV (2017)

\bibitem{goodfellow2016dlbook}
Goodfellow, I., Bengio, Y., Courville, A.: Deep Learning. MIT Press (2016)

\bibitem{goodfellow2014explaining}
Goodfellow, I.J., Shlens, J., Szegedy, C.: Explaining and harnessing
  adversarial examples. ICLR  (2015)

\bibitem{Guyon:IJCNN:2015}
Guyon, I., Bennett, K., Cawley, G., Escalante, H.J., Escalera, S., Ho, T.K.,
  Maci\`{a}, N., Ray, B., Saeed, M., Statnikov, A., Viegas, E.: Design of the
  2015 {ChaLearn} {AutoML} challenge. In: IJCNN (2015)

\bibitem{Guyon:AutoML:2016}
Guyon, I., Chaabane, I., Escalante, H.J., Escalera, S., Jajetic, D., Lloyd,
  J.R., Mac\'ia, N., Ray, B., Romaszko, L., Sebag, M., Statnikov, A., Treguer,
  S., Viegas, E.: A brief review of the {ChaLearn} {AutoML} challenge. In:
  AutoML workshop@ICML (2016)

\bibitem{he2016deep}
He, K., Zhang, X., Ren, S., Sun, J.: Deep residual learning for image
  recognition. In: CVPR (2016)

\bibitem{irpan2018deeprl}
Irpan, A.: Deep reinforcement learning doesn't work yet. Online (Feb. 14):
  \url{https://www.alexirpan.com/2018/02/14/rl-hard.html} (2018)

\bibitem{DBLP:journals/corr/JohnL17}
John, V., Liu, X.: A survey of distributed message broker queues. arXiv
  preprint arXiv:1704.00411  (2017)

\bibitem{krizhevsky2012imagenet}
Krizhevsky, A., Sutskever, I., Hinton, G.E.: {ImageNet} classification with
  deep convolutional neural networks. In: NIPS (2012)

\bibitem{larochelle2009exploring}
Larochelle, H., Bengio, Y., Louradour, J., Lamblin, P.: Exploring strategies
  for training deep neural networks. JMLR (1),  1--40 (1 2009)

\bibitem{lecun1998efficient}
LeCun, Y., Bottou, L., Orr, G.B., Müller, K.R.: Efficient backprop. In: Orr,
  G.B., Müller, K.R. (eds.) Neural networks: Tricks of the trade, pp. 9--50.
  Springer, Berlin, Heidelberg (1998)

\bibitem{li_wang_tan_jain_2004}
Li, J., Wang, Y., Tan, T., Jain, A.K.: Live face detection based on the
  analysis of {Fourier} spectra. Biometric Technology for Human Identification
  (2004)

\bibitem{DBLP:conf/ipta/LiFBXLH16}
Li, L., Feng, X., Boulkenafet, Z., Xia, Z., Li, M., Hadid, A.: An original face
  anti-spoofing approach using partial convolutional neural network. In: Int.
  Conference on Image Processing Theory, Tools and Applications (IPTA) (2016)

\bibitem{liu2017survey}
Liu, W., Wang, Z., Liu, X., Zeng, N., Liu, Y., Alsaadi, F.E.: A survey of deep
  neural network architectures and their applications. Neurocomputing
  \textbf{234},  11 -- 26 (2017)

\bibitem{lowe2017multi}
Lowe, R., Wu, Y., Tamar, A., Harb, J., Abbeel, O.P., Mordatch, I.: Multi-agent
  actor-critic for mixed cooperative-competitive environments. In: NIPS (2017)

\bibitem{DBLP:conf/icb/MaattaHP11}
M{\"{a}}{\"{a}}tt{\"{a}}, J., Hadid, A., Pietik{\"{a}}inen, M.: Face spoofing
  detection from single images using micro-texture analysis. In: Int. Joint
  Conference on Biometrics (IJCB) (2011)

\bibitem{meier2017fully}
Meier, B., Stadelmann, T., Stampfli, J., Arnold, M., Cieliebak, M.: Fully
  convolutional neural networks for newspaper article segmentation. In: ICDAR
  (2017)

\bibitem{mnih2013playing}
Mnih, V., Kavukcuoglu, K., Silver, D., Graves, A., Antonoglou, I., Wierstra,
  D., Riedmiller, M.: Playing {Atari} with deep reinforcement learning. arXiv
  preprint arXiv:1312.5602  (2013)

\bibitem{ng2018yearning}
Ng, A.: Machine Learning Yearning - Technical Strategy for AI Engineers in the
  Era of Deep Learning (2018), [to appear]

\bibitem{olah2017research}
Olah, C., Carter, S.: Research debt. Distill  (2017)

\bibitem{olah2018buildingblocks}
Olah, C., Satyanarayan, A., Johnson, I., Carter, S., Schubert, L., Ye, K.,
  Mordvintsev, A.: The building blocks of interpretability. Distill  (2018)

\bibitem{Olson2016EvoBio}
Olson, R.S., Urbanowicz, R.J., Andrews, P.C., Lavender, N.A., Kidd, L.C.,
  Moore, J.H.: Automating biomedical data science through tree-based pipeline
  optimization. In: European Conference on the Applications of Evolutionary
  Computation (EvoApplications) (2016)

\bibitem{pan2010survey}
Pan, S.J., Yang, Q.: A survey on transfer learning. IEEE Trans. Knowledge and
  Data Engineering  \textbf{22}(10),  1345--1359 (2010)

\bibitem{vggface2015}
Parkhi, O.M., Vedaldi, A., Zisserman, A.: Deep face recognition. In: BMVC
  (2015)

\bibitem{DBLP:journals/tifs/PatelHJ16}
Patel, K., Han, H., Jain, A.K.: Secure face unlock: Spoof detection on
  smartphones. {IEEE} Trans. Information Forensics and Security
  \textbf{11}(10),  2268--2283 (2016)

\bibitem{perez2017playbook}
Perez, C.E.: The Deep Learning AI Playbook - Strategy for Disruptive Artificial
  Intelligence (2017)

\bibitem{rajpurkar2017mura}
Rajpurkar, P., Irvin, J., Bagul, A., Ding, D., Duan, T., Mehta, H., Yang, B.,
  Zhu, K., Laird, D., Ball, R.L., et~al.: {MURA} dataset: Towards
  radiologist-level abnormality detection in musculoskeletal radiographs. arXiv
  preprint arXiv:1712.06957  (2017)

\bibitem{DBLP:journals/ijmir/RebeloFPMGC12}
Rebelo, A., Fujinaga, I., Paszkiewicz, F., Mar{\c{c}}al, A.R.S., Guedes, C.,
  Cardoso, J.S.: Optical music recognition: state-of-the-art and open issues.
  Int. Journal of Multimedia Information Retrieval  \textbf{1}(3),  173--190
  (2012)

\bibitem{DBLP:conf/nips/RenHGS15}
Ren, S., He, K., Girshick, R.B., Sun, J.: Faster {R-CNN:} towards real-time
  object detection with region proposal networks. In: NIPS (2015)

\bibitem{ILSVRC15}
Russakovsky, O., Deng, J., Su, H., Krause, J., Satheesh, S., Ma, S., Huang, Z.,
  Karpathy, A., Khosla, A., Bernstein, M., Berg, A.C., Fei-Fei, L.: {ImageNet
  Large Scale Visual Recognition Challenge}. Int. Journal of Computer Vision
  \textbf{115}(3),  211--252 (2015)

\bibitem{schmidhuber2015deep}
Schmidhuber, J.: Deep learning in neural networks: An overview. Neural networks
   \textbf{61},  85--117 (2015)

\bibitem{schroff2015facenet}
Schroff, F., Kalenichenko, D., Philbin, J.: {FaceNet}: A unified embedding for
  face recognition and clustering. In: CVPR (2015)

\bibitem{schulman2017proximal}
Schulman, J., Wolski, F., Dhariwal, P., Radford, A., Klimov, O.: Proximal
  policy optimization algorithms. arXiv preprint arXiv:1707.06347  (2017)

\bibitem{shalev2012online}
Shalev-Shwartz, S.: Online learning and online convex optimization. Foundations
  and Trends in Machine Learning  \textbf{4}(2),  107--194 (2012)

\bibitem{shwartz2017opening}
Shwartz{-}Ziv, R., Tishby, N.: Opening the black box of deep neural networks
  via information. arXiv preprint arXiv:1703.00810  (2017)

\bibitem{silver2016mastering}
Silver, D., Huang, A., Maddison, C.J., Guez, A., Sifre, L., Van Den~Driessche,
  G., Schrittwieser, J., Antonoglou, I., Panneershelvam, V., Lanctot, M.,
  et~al.: Mastering the game of {Go} with deep neural networks and tree search.
  nature  \textbf{529}(7587),  484--489 (2016)

\bibitem{simonyan2014very}
Simonyan, K., Zisserman, A.: Very deep convolutional networks for large-scale
  image recognition. ICLR  (2015)

\bibitem{springenberg2014striving}
Springenberg, J.T., Dosovitskiy, A., Brox, T., Riedmiller, M.: Striving for
  simplicity: The all convolutional net. arXiv preprint arXiv:1412.6806  (2014)

\bibitem{stadelmann2018beyondimagenet}
Stadelmann, T., Tolkachev, V., Sick, B., Stampfli, J., D\"urr, O.: Beyond
  {ImageNet} - deep learning in industrial practice. In: Braschler, M.,
  Stadelmann, T., Stockinger, K. (eds.) Applied Data Science - Lessons Learned
  for the Data-Driven Business. Springer (2018), [to appear]

\bibitem{sutskever2013importance}
Sutskever, I., Martens, J., Dahl, G., Hinton, G.: On the importance of
  initialization and momentum in deep learning. In: ICML (2013)

\bibitem{szegedy2015going}
Szegedy, C., Liu, W., Jia, Y., Sermanet, P., Reed, S., Anguelov, D., Erhan, D.,
  Vanhoucke, V., Rabinovich, A., et~al.: Going deeper with convolutions. CVPR
  (2015)

\bibitem{tuggener2018deepscores}
Tuggener, L., Elezi, I., Schmidhuber, J., Pelillo, M., Stadelmann, T.:
  {DeepScores} - a dataset for segmentation, detection and classification of
  tiny objects. In: ICPR (2018)

\bibitem{DBLP:conference/ismir/tuggener}
Tuggener, L., Elezi, I., Schmidhuber, J., Stadelmann, T.: Deep watershed
  detector for music object recognition. In: ISMIR (2018)

\bibitem{xu2017performance}
Xu, P., Shi, S., Chu, X.: Performance evaluation of deep learning tools in
  {Docker} containers. arXiv preprint arXiv:1711.03386  (2017)

\bibitem{DBLP:conf/acpr/XuLD15}
Xu, Z., Li, S., Deng, W.: Learning temporal features using {LSTM-CNN}
  architecture for face anti-spoofing. In: ACPR (2015)

\bibitem{DBLP:journals/corr/YangLL14}
Yang, J., Lei, Z., Li, S.Z.: Learn convolutional neural network for face
  anti-spoofing. arXiv preprint arXiv:1408.5601  (2014)

\bibitem{DBLP:conf/nips/YosinskiCBL14}
Yosinski, J., Clune, J., Bengio, Y., Lipson, H.: How transferable are features
  in deep neural networks? In: NIPS (2014)

\bibitem{DBLP:conf/icb/ZhangYLLYL12}
Zhang, Z., Yan, J., Liu, S., Lei, Z., Yi, D., Li, S.Z.: A face antispoofing
  database with diverse attacks. In: Int. Conference on Biometrics (ICB) (2012)

\bibitem{zheng2016improving}
Zheng, S., Song, Y., Leung, T., Goodfellow, I.: Improving the robustness of
  deep neural networks via stability training. In: CVPR (2016)

\end{thebibliography}
\bibliographystyle{splncs04}

\end{document}